\documentclass{svproc}

\usepackage[T1]{fontenc}                  
\usepackage[utf8]{inputenc}               

\usepackage{graphicx}                     
\usepackage{multicol}                     
\usepackage{footmisc}                     

\usepackage{algorithm}
\usepackage{algorithmic}
\usepackage{amssymb,amsmath}
\usepackage{graphics}
\usepackage[usernames, dvipsnames]{xcolor}

\usepackage{booktabs}
\usepackage{multirow}
\usepackage{subcaption}
\newcommand{\NA}{\textemdash}

\begin{document}

\title{Annotated-skeleton Biased Motion Planning for Faster Relevant Region
Discovery}

\titlerunning{Annotated Skeleton Biased Planning}

\author{Diane Uwacu\inst{1}
\and Regina Rex\inst{2}
\and Bonnie Wang\inst{3}
\and Shawna Thomas\inst{1}
\and Nancy M. Amato\inst{4}}

\authorrunning{Diane Uwacu et al.}

\tocauthor{Diane Uwacu,
Regina Rex,
Bonnie Wang,
Shawna Thomas,
Nancy M. Amato}

\institute{%
Parasol Laboratory,
Texas A\&M University Dept. of Computer Science and Engineering,
College Station, TX, 77840, USA.\\
\email{duwacu@tamu.edu}, \email {sthomas@cse.tamu.edu} \texttt{http://parasol.tamu.edu}
\and
University of Wisconsin-Superior Dept. of Mathematics and Computer Science,
Superior, WI, 54880, USA.\\
\email{rrex@uwsuper.edu}.
\and
Columbia University Dept. of Computer Science,
New York, NY, 10027, USA.\\
\email{bw2551@columbia.edu}.
\and
Parasol Laboratory,
University of Illinois Urbana-Champaign Dept. of Computer Science,
Champaign, IL, 61820, USA.\\
\email{namato@illinois.edu}
}

\maketitle

\begin{abstract}

Motion planning algorithms often leverage topological information about the
environment to improve planner performance.
However, these methods often focus only on the environment's connectivity while
ignoring other properties such as obstacle
clearance, terrain conditions and resource accessibility.
We present a method that augments a skeleton representing the workspace topology with
such information to guide a sampling-based motion planner
to rapidly discover regions most relevant to the problem at hand.
Our approach decouples guidance and planning, making it possible for
basic planning algorithms to find desired paths earlier in the planning process.
We demonstrate the efficacy of our approach in both robotics problems and
applications in drug design.
Our method is able to produce desirable paths quickly with no change to the
underlying planner.
\end{abstract}

\section{Introduction}
Motion planning, finding a valid path for a movable object from
one state to another, has applications in many domains including robotics, computer-aided
design, and computational biology \cite{l-mpjrmdaoa-99}.
Problems in these applications are usually defined with conditions on the quality
of the solution.
For example, path safety is paramount to robot navigation, availability of recharging
stations is important for mission completion in robotics, and energy feasibility is
critical in protein-ligand binding simulations.
While motion planning algorithms can successfully handle
complex motion planning problems, balancing fast computation and
additional path quality conditions is a challenge.\par

To address the fast computation part, one can take
advantage of topological information when solving problems whose
workspace and configuration space are closely related.
In such scenarios, the workspace topology knowledge can be used to identify regions of the environment and their connectivity, to be explored by the planner \cite{dsba-drbrrt-16,rhb-iotpodt-2017,rmft-otppsrn-2017,hpc-tpiasocce-2019}.
This guidance can increase the speed of  planning process.
In applications that require a certain quality of the path however, the planner,
even guided topologically, can spend time exploring parts of the environment
that are not relevant to the needed solution.\par
Prior research in path quality often focuses on one metric. Planning methods
like medial axis planning \cite{lta-gfsmafs-03,yb-asdppbama-04} push every
configuration sampled to the medial axis.
This provides paths with high clearance from obstacles, which is helpful in
scenarios like the one mentioned above, where path safety is needed for navigation.
However, metric-specific methods like medial axis planning cannot be as easily
generalized for other metrics like region priority. \par

In this work, we present annotated-skeleton biased planning, a method that
embeds environment information in a topological skeleton to guide sampling-based
motion planning algorithms. Given a biasing metric, the method annotates
the workspace skeleton with its related information, and the planner explores the
environment along the skeleton, biasing towards the specified requirement.
\par
We combine
workspace guidance and environment-specific information to steer planning.
Our method presents an advantage over metric-specific planning algorithms that are
biased towards a single metric, by separating planning from biasing, making it possible to use any underlying
planner to achieve the specified path quality requirements.\par

This method is tested on robotics environments and protein-ligand binding
applications using different biases like path safety and binding energy to demonstrate its
generality and flexibility.
The method is compared to a skeleton guided planner
and achieves the desired planning behavior with a negligible time trade-off.\par

\section{Related Work}
As the environment gets more complex or as the degrees of freedom of the movable object increase, exact motion planning solutions
become intractable \cite{r-cpmpg-79}, hence the popularity of sampling-based
algorithms that use randomness and heuristics to solve high-dimensional complex
problems.
Sampling-based algorithms can be subdivided into two categories, namely, the tree-based rapidly exploring
random trees (RRT) \cite{lk-rkp-01} and the graph-based probabilistic roadmaps (PRM) \cite{kslo-prpp-96}.
Typically, RRT-based algorithms optimize speed and are easily applied to robotics
problems with nonholonomic constraints, while PRM-based algorithms
optimize connectivity and can be applied to multi-query problems.
Rapidly exploring random graphs (RRG) \cite{r-rrgmpmmr}
are a hybrid of both algorithms that combines their strengths.
There have been variants of heuristics applied to these algorithms to improve
planning in specific conditions.

\subsection{Topology-Guided Planning}
Topological guidance involves using the workspace structure to direct how sampling-based planners explore the environment.
Some motion planning approaches use the workspace decomposition for planning,
especially to target narrow passages.
For example, Vonasek et al \cite{vk-tdpssmp-2017} use Voronoi diagrams to guide
an RRT through a protein environment and discover possible protein tunnels that
could connect to a binding site. \par

Several topology-guided planners use workspace skeletons, which are graphs that
capture the topological features of the environment.
One such planner is Dynamic Region-biased RRT (DR-RRT) \cite{dsba-drbrrt-16}.
This strategy guides the planner to sample only from particular regions based on
the workspace topology.
A skeleton, or graph that maps the essence of the free workspace topology,
directs region selection.
These regions are created, sampled based on their previous exploration success
rate, and then destroyed as the planner explores new regions.

Topological guidance has been used for online planning
\cite{rhb-iotpodt-2017,rmft-otppsrn-2017}, and in control problems for planning
with dynamic environments with interactions \cite{vdmt-trmpgdei-2013}. Ha et al
\cite{hpc-tpiasocce-2019} recently defined H-signatures to find optimal paths in cluttered environments.
\par

\subsection{Property-Specific Planning}
Property-specific planners guide the planning process with objective-based
functions that depend on certain environment features. Such planners target
specific environment characteristics relevant to the path requirements.
For example, medial axis based algorithms \cite{lta-gfsmafs-03,yb-asdppbama-04}
aim to build a roadmap along the medial axis of the environment, which
results in paths with maximized obstacle clearance.
On the other hand, obstacle based algorithms
\cite{abdjv-obprm-98,bsa-lbobprm-01-book,ytea-uudobp-12} increase chances of
finding shortest paths through narrow passages by building a roadmap close to
obstacle surfaces.
Although medial axis and obstacle based algorithms successfully return a roadmap
with the desired properties, they can be computationally expensive because the planner
has to interact with the workspace to guide planning.
In addition, these methods and others \cite{yl-fpfmsbrng-00,hk-fuwmapp-00} that are
property-specific are not capable of handling other priorities. \par

\section{Methodology}
\begin{figure*}[!htbp]
  \centering
\begin{subfigure}{.45\textwidth}
  \centering
  \includegraphics[width=1\textwidth]{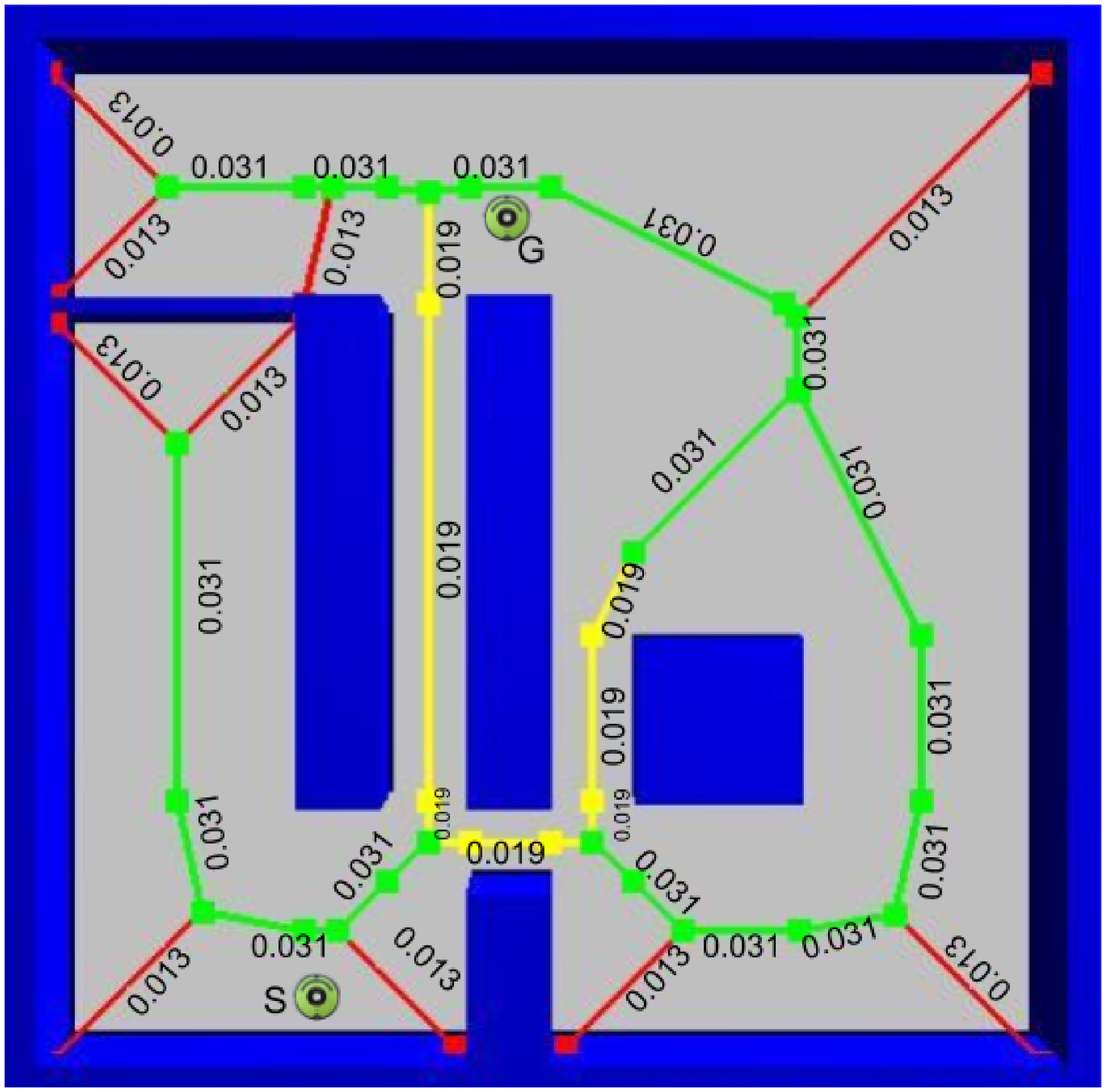}
  \caption{}
  \label{fig:annotated_skeleton}
\end{subfigure}
  \centering
\begin{subfigure}{.45\textwidth}
  \centering
  \includegraphics[width=1\textwidth]{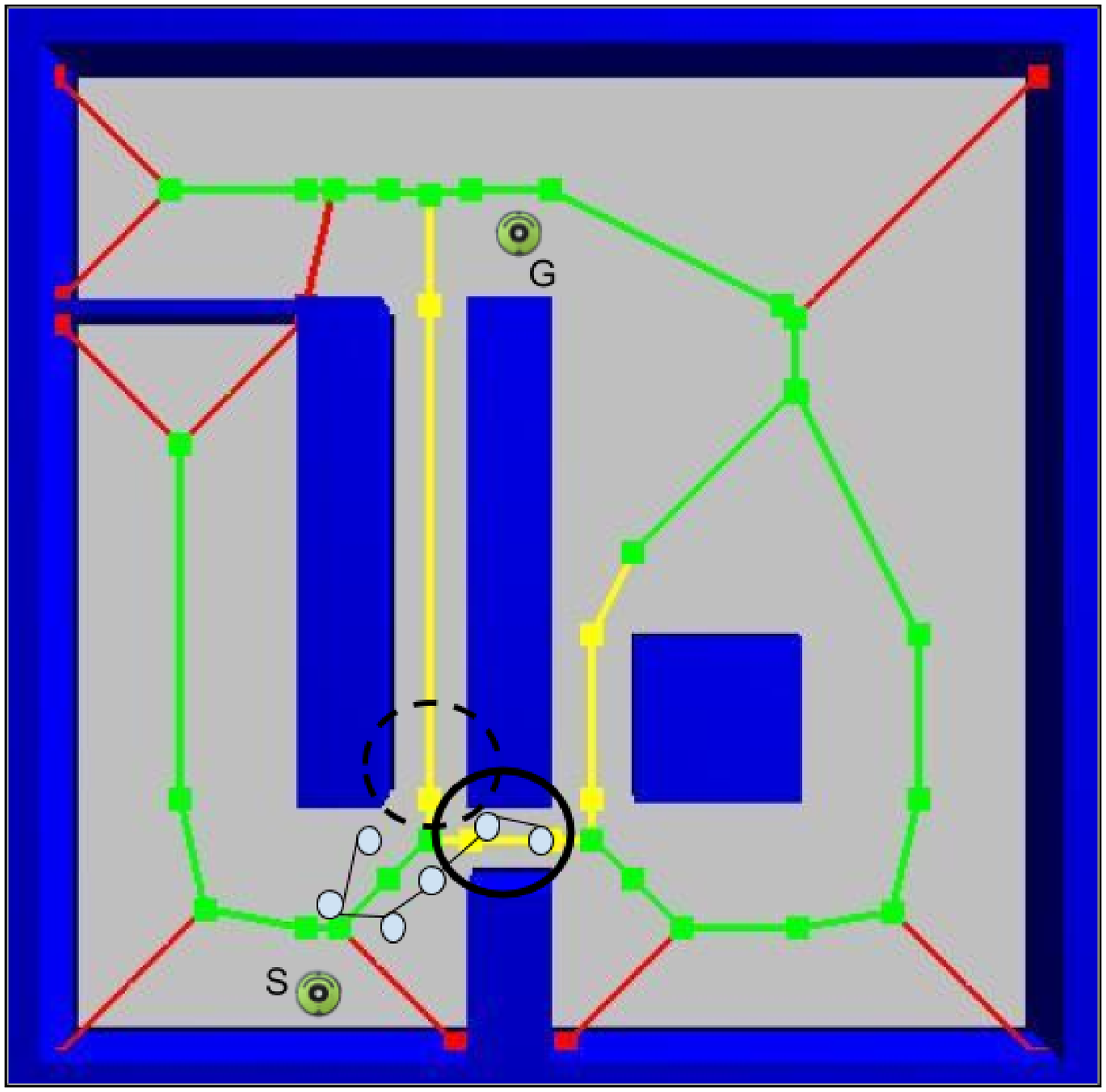}
  \caption{}
  \label{fig:guided_planning}
\end{subfigure}
  \centering
\begin{subfigure}{.45\textwidth}
  \centering
  \includegraphics[width=1\textwidth]{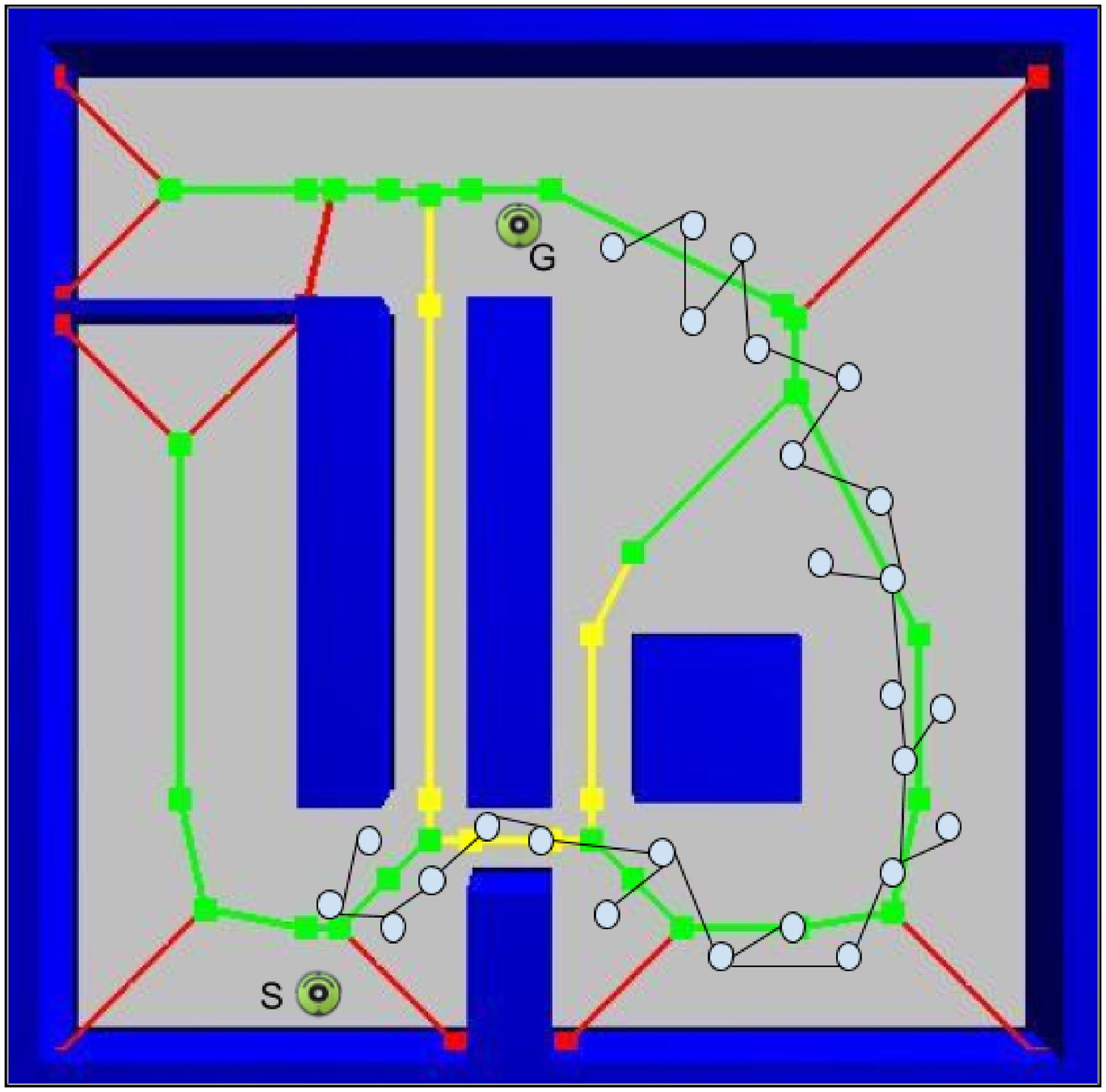}
  \caption{}
  \label{fig:max_clearance_path}
\end{subfigure}
  \centering
\begin{subfigure}{.45\textwidth}
  \centering
  \includegraphics[width=1\textwidth]{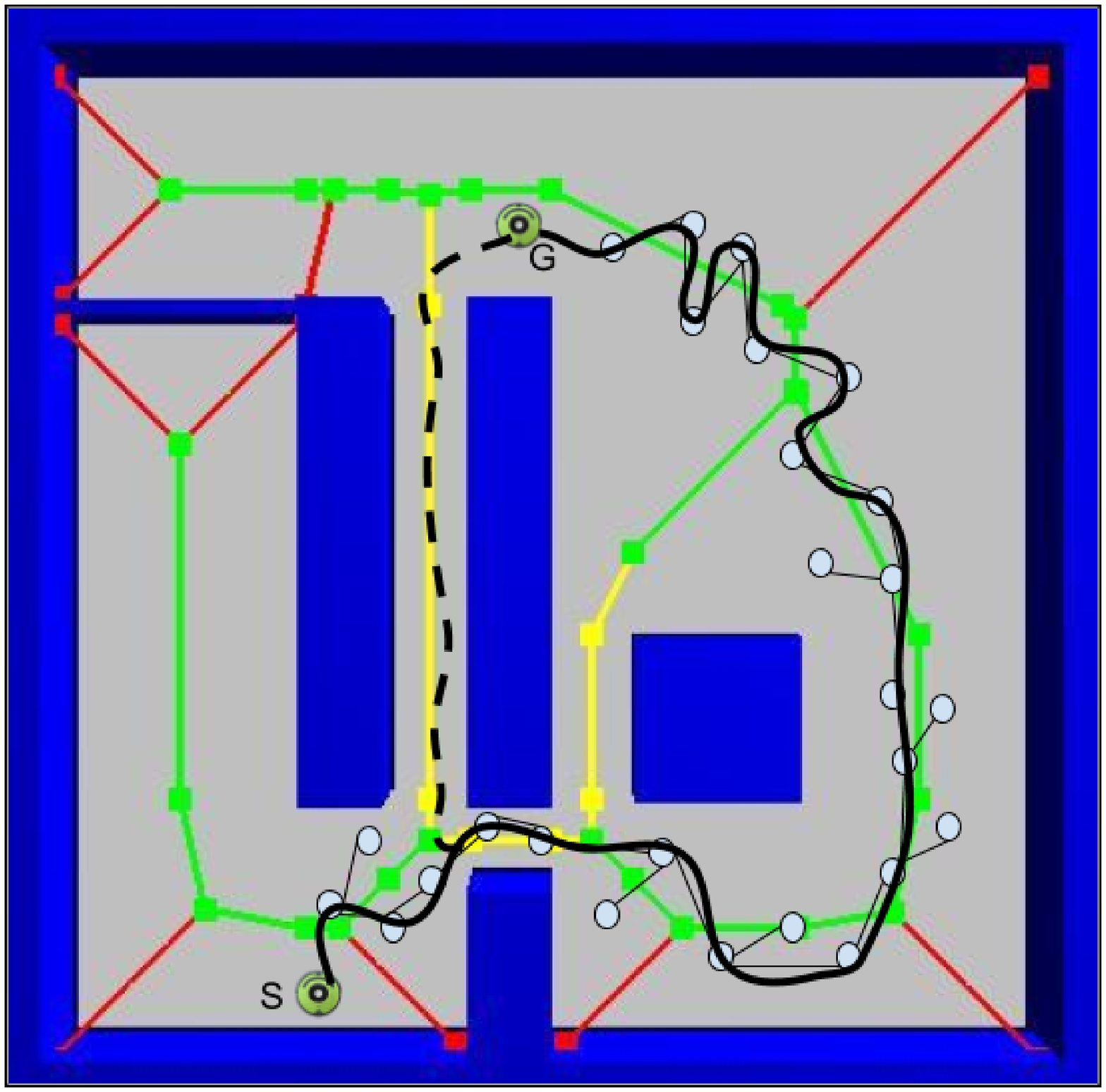}
  \caption{}
  \label{fig:max_clearance_path}
\end{subfigure}
\caption{Example execution of the annotated-skeleton biased method: (a)
  Workspace Skeleton annotated with clearance value (red $\rightarrow$ lowest
  clearance, yellow $\rightarrow$ medium clearance, green $\rightarrow$ highest clearance). The query is shown as S for
  start and G for goal. (b) Region selection based on maximum clearance-bias. A
  region over a shorter edge is preferred to a longer one in poor conditions.
  (c) Exploration process where the passage on the right is explored first.
  (d) Extracted maximum-clearance path. Shorter less safe alternative is shown
in dashed line.}
 \label{fig:MethodExplanation}
\end{figure*}

The annotated-skeleton biased planning algorithm annotates the workspace
skeleton with biasing metric information and then guides the planner based on
this information to increase chances of finding desirable paths faster.
In the following sections we explain how the skeleton is constructed and annotated with
environment properties and how regions of the environment are dynamically selected to bias
the discovery of regions following a given metric.
Algorithm \ref{alg:alg1} outlines the annotated-skeleton biased approach, and Figure
\ref{fig:MethodExplanation} illustrates a simple example of a planner guided by
a clearance-based annotated skeleton.

\subsection{Constructing and Annotating the Skeleton}
Given a motion planning problem and some biasing metric, we
first construct a skeleton through the free workspace.
A workspace skeleton is a unit-dimensional curve fully contained inside the
environment free space such that the free space can be collapsed into the
skeleton in a continuous way \cite{blk-tcsrp-2012}.
There are different ways of constructing the workspace skeleton, and for our
purpose, we are usually most interested in a medially centered structure so that we can
accurately extract local properties of the environment.
For that reason, we construct a Medial Axis \cite{c-edt-92} in 2D environments
and use a Mean Curvature Skeleton \cite{taos-mcs-12} in 3D environments.\par

We annotate the skeleton with environment properties by saving for each edge the
bottleneck w.r.t. each property.
This is so that we can get the property value that can inform us on the highest cost to incur in the region represented by the edge.
If we are interested in path safety in relation to obstacle clearance for
example, we save the lowest obstacle clearance of the
skeleton edge. In a similar way, for the protein-ligand binding application,
where we need to guide the planner towards high volume and low energy regions,
we save the highest energy along the edge in addition to the lowest obstacle
clearance value.
A clearance-annotated skeleton can be seen in Figure
\ref{fig:annotated_skeleton}. It is color-coded from highest to lowest clearance, starting
with green to yellow and ending with red.\par
The annotated skeleton is robot-independent and can be reused in the same
workspace with different types of robots.
We demonstrate this in the protein-ligand application by using a set of ligand
probes, commonly used in protein-ligand binding site prediction research
\cite{lj-qemppbs-05}, to
compute van der Waals attraction forces at different
protein locations along the mean curvature skeleton edges.

\begin{algorithm}[h]
 \caption{Annotated-skeleton biased planner}
 \label{alg:alg1}
 \begin{algorithmic}[1]
 \renewcommand{\algorithmicrequire}{\textbf{Input:}}
 \renewcommand{\algorithmicensure}{\textbf{Output:}}
 \REQUIRE Environment $env$, Start $s$, Goal $g$, Bias Metric $\mu$
 \ENSURE Path $P$
  \STATE{$WS \leftarrow {\tt  GetWorkspaceSkeleton}(env)$}
  \STATE{$AS \leftarrow {\tt  AnnotateSkeleton}(env, WS, env.Properties)$}
  \STATE{Initialize $R$ to $\phi$}
  \STATE{$v \leftarrow {\tt  GetInitialVertex}(start, AS)$}
  \STATE{$\{A_{r}\} \leftarrow {\tt CreateActiveRegions}(v)$}
  \WHILE{$\lnot done$}
  \STATE{$r \leftarrow {\tt SelectRegion}(\{A_{r}\}$, $\mu)$}
  \IF{$r == NULL$}
  \STATE{$r \leftarrow env.{\tt GetBoundary}()$}
  \ENDIF
  \STATE{$R \leftarrow {\tt GrowRoadmap}(r, \mu)$}
  \STATE{${\tt AdvanceRegion}(r)$}
  \IF{${\tt RegionReachedEndOfEdge}(G, r)$}
  \STATE{$v \leftarrow {\tt GetNextVertex}(r)$}
  \STATE{$\{A_{r}\} \leftarrow \{A_{r}\} \setminus r$}
  \STATE{$\{A_{r}\} \leftarrow {\tt CreateActiveRegions}(v)$}
  \ENDIF
  \ENDWHILE
  \STATE{$P \leftarrow {\tt Query}(R, s, g)$}
 \RETURN $P$
 \end{algorithmic}
 \end{algorithm}

\subsection{Growing the Roadmap with Biasing}
In this phase we grow a sampling-based roadmap that explores environment regions
that are most relevant to the biasing metric.
We start by finding the closest skeleton vertex to the start
configuration as shown in Line 4 of the algorithm.
We then spark a spherical sampling regions along each outgoing edge of the vertex. These regions have a
fixed radius equivalent to the size of the robot.\par
The following steps from Line 6 to 18 are repeated until the problem is solved.
A sampling region is selected from the set of active regions based on the
annotation of the skeleton edge along which the region is defined.
The selection is done in three steps shown in Algorithm \ref{alg:alg3}.
Each active region is weighted based on the skeleton edge annotation.
The relative edge length is used as a scaling factor to take into consideration how much time it would take to traverse the edge.
As illustrated in Figure \ref{fig:guided_planning}, factoring the
relative length of the edge into the selection process allows the planner to make a
choice between two equally poorly qualified regions based on the expected
duration of the poor conditions.\par

\begin{algorithm}[h]
 \caption{SelectRegion}
 \label{alg:alg3}
 \begin{algorithmic}[1]
 \renewcommand{\algorithmicrequire}{\textbf{Input:}}
 \renewcommand{\algorithmicensure}{\textbf{Output:}}
 \REQUIRE active regions set $\{R\}$, biasing metric $\mu$
 \ENSURE active region $r$
 \STATE{$r \leftarrow NULL$}
 \FOR{each $r_i \in \{R\}$}
 \STATE{$a \leftarrow r_i.{\tt GetSkeletonEdge}().{\tt
 GetMetric}(\mu)$}
 \STATE{$l \leftarrow r_i.{\tt GetSkeletonEdge}().{\tt GetLength}()$}
 \STATE{$region\_weights.{\tt push\_back}({\tt
 Weight}(a^{l/minL}))$}
 \ENDFOR
 \STATE{$r \leftarrow {\tt min\_arg}(region\_weights)$}
 \RETURN $r$
 \end{algorithmic}
 \end{algorithm}

The roadmap is extended towards the selected region using an RRT Extend step.
Depending on the problem, ${\tt GrowRoadmap}()$ builds a tree or a graph. In
applications where we are interested in increasing the quality of the returned
path, such as the protein-ligand binding application, we increase connections and
build an RRG.\par
These steps increase the chance of the first sequence of explored regions to
contain a path that satisfies the given conditions.
At each iteration, the algorithm greedily attempts to expand the roadmap towards
regions with a higher probability of satisfying the query conditions.\par

After the roadmap is constructed, it is queried to find the paths to the
goal using a graph search algorithm.
The quality of the extracted path is dependent in part on the density of the
roadmap. This in turn is dependent in part on the sampling region size as well as the
step size for the RRT extension or the PRM local planner move.
To save computation time, we use fixed-size spherical sampling regions defined based on
the size of the robot, and in the Query step in Line 19 of Algorithm \ref{alg:alg1}, we find homotopic paths to the extracted one using smoothing algorithms.
.\par

\subsection{Probabilistic Completeness}
Given that the workspace skeleton represents the free space, a skeleton-guided planner with unlimited resources, operating in a region of radius $r$ big enough to cover the whole environment is probabilistically complete.
For that reason, as proposed by Denny et al \cite{dsba-drbrrt-16}, we keep a probability $p > 0$ of selecting the full environment as an active sampling region in order to conserve completeness.\par
In addition, to the theoretical completeness, the method is experimentally faster than traditional unguided planning methods. This can be explained by two properties of topological guidance:
First, the workspace skeleton indicates the connectivity of the free space, making it possible to find narrow passages as easily as wide areas are found. Second, region sizes are conservatively set so that they contain a valid configuration of the robot while minimizing chances of covering obstacle space. This is done by using a medially centered skeleton and a spherical region of radius $r = robot.Length$.\par
Annotating the skeleton with information relevant to the problem similarly indicates regions that are important to the solution, giving the planner a higher chance of visiting them earlier.\par

\section{Experiments}
We used robotics and computational biology examples to study the effect of using
an annotated skeleton to guide a planner through different definitions of quality.
In robotics environments, we used maximum clearance to bias towards safer
regions.
In protein environments, we used two different metrics: maximum clearance to find high volume
protein tunnels and minimum energy to find low energy tunnels.\par
Our experiments test the ability of using a general planner to achieve different specific
planning behaviors, which is impossible to do with any one specialized planner.

\subsection{Environments studied}
We studied two robotics examples and two protein variants to study protein-ligand binding.
\subsubsection{Robotics Environments}
To analyze the general applicability of our method, we tested 2D and 3D robotics environment with robots of varying constraints.
Both environments have more than one possible path to solve the query. We explicitly made the most desirable path different from the easily discoverable one to test the ability of the planner to find the desired path.\par
We define success as the ability to return paths relevant to the query
conditions.
After planning, we analyzed paths checking the ones that went through a sequence
of regions that maximizes clearance.
\begin{itemize}
  \item \textbf{3D Boxes}(Fig \ref{fig:boxes_results}): This is a 6 DOF  rigid body robot in an environment
      with three boxes. This 3D example was used to test the efficiency of mean curvature skeleton
      guidance.
    \item \textbf{Hallways}(Fig \ref{fig:walls_results}): A 3 DOF nonholonomic
      iRobot Create is tested in a planar environment with hallways of different
      widths. Safe navigation is important in this environment to allow a margin
      of error when the path found is deployed on a physical system.
\end{itemize}

\subsubsection{Protein-Ligand Binding}
\par
A protein is a large molecular structure made from a chain of amino acids,
involved in several essential metabolic reactions.
It reacts with a drug molecule called a ligand to activate or inhibit its
functionality, through a process called protein-ligand binding. This process is conditioned by
geometric and energetic compatibility between the protein's molecules and the ligand.
Molecules involved in the binding process form the protein's binding site. A
large number of proteins have their binding sites buried inside the cavity of
the protein and accessible from the outside through protein tunnels.
Molecules that line up those tunnels often regulate the binding site's accessibility,
further increasing the protein's selectivity \cite{kaushik2018}.\par

 The protein configuration data is procured from the Protein Data Bank (PDB)
 \cite{PDB-webpage} and a geometric structure can be constructed using a
 software called CHIMERA \cite{chimera2004}. We model the ligand as a 3D
 rotational multi-linkage robot with a torsional joint degree of freedom per
 linkage.\par
Given a protein geometry as an environment, we use motion planning to find
possible tunnels inside the protein.
The start configuration is generated by sampling a valid configuration of the
ligand with maximum contact with the protein binding site molecules.
Using the workspace skeleton for guidance, we explore tunnels that connect the
binding site to the outside surface. An annotated skeleton can help explore
regions inside the protein that are likely to contain viable tunnels. These are
regions with low energy cost and high obstacle clearance.\par
After planning, we analyze paths found by
classifying them using a scoring function to evaluate tunnel accessibility. The
scoring function takes into consideration the tunnel volume, energy minima, and
bottlenecks. \par

We studied haloalkane dehalogenase (DhaA) protein (Figure
\ref{fig:4hzg_path}). It is used in the degradation of soil pollutants
like tri-chloropropane (TCP). DhaA binds with TCP and the velocity of
that reaction depends on the accessibility of DhaA's binding site. To increase
its reactivity, biochemists have engineered DhaA mutants that are more
accessible to the ligand. We used motion planning to compare the native
structure (wild type) and one of its mutants, DhaA31.
The success ratio was determined by the number of
tunnels discovered compared to the experimentally known number of tunnels for
the protein-ligand pair.
\begin{itemize}
    \item \textbf{DhaA}: The active site of DhaA is
      connected to the protein surface
by two major access tunnels, although three more tunnels have been predicted
through molecular modeling \cite{KLVANA20091339}.
    \item  \textbf{DhaA31}: To increase its activity
      towards TCP, certain residues were modified, which reduced the size of the
      binding pocket to fit the ligand and leave little space to the solvent.
      This was reported to increase DhaA’s activity towards TCP. Two tunnels are
      documented for this protein \cite{Lahoda:kw5074}.
\end{itemize}

\subsection{Experimental Setup}
We run three types of strategies and analyze the planner's performance in terms of speed
and ability to satisfy path requirements, and the role that guidance plays in
that:
\begin{itemize}
    \item \textbf{MA-RRT/MA-RRG}: Medial axis RRT/RRG without topological
      guidance. This is an example of a specialized planner, expected to find
      paths near the medial axis of the environment, maximizing obstacle clearance.
    \item \textbf{DR-RRT/DR-RRG}: Dynamic Region-biased RRT/RRG with topological
      guidance without environment property annotation. It is expected to
      improve planning speed by guiding the planner away from workspace regions
      occupied by obstacles.
    \item \textbf{AB-RRT/AB-RRG}: Our method, Annotated-skeleton Biased RRT/RRG guided by an
      annotated workspace skeleton.
\end{itemize}

The robotics environments were run with 40 seeds each while the protein
experiments were run with 10 seeds. MA-RRG is not reported in the protein
environments because it could not solve the problem within the time limit of four minutes.

\subsection{Results and Discussion}
Table \ref{tab:runtime} shows the running times for the robotics environments.
In the skeleton construction phase, we observed that the overhead added by the
annotation was less than 0.01 seconds. Note that MA-RRT does not rely on a
workspace skeleton and incurs the cost of computing a medial axis during run
time.\par
\begin{figure*}[htb]
  \centering
  \begin{subfigure}{0.45\textwidth}
  \centering
  \includegraphics[width=1\textwidth]{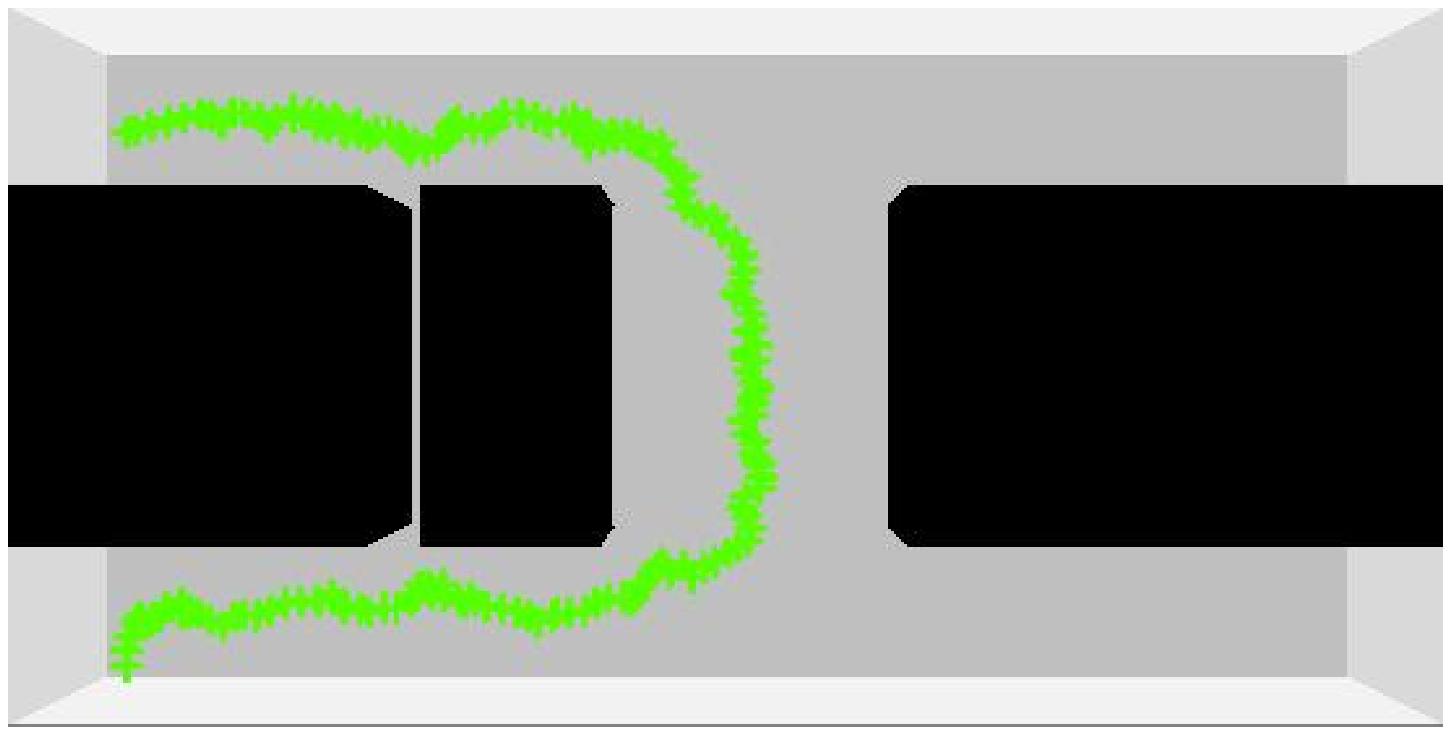}
  \caption{AB-RRT}
  \label{fig:abrrt_boxes}
\end{subfigure}
  \centering
  \begin{subfigure}{0.45\textwidth}
  \centering
  \includegraphics[width=1\textwidth]{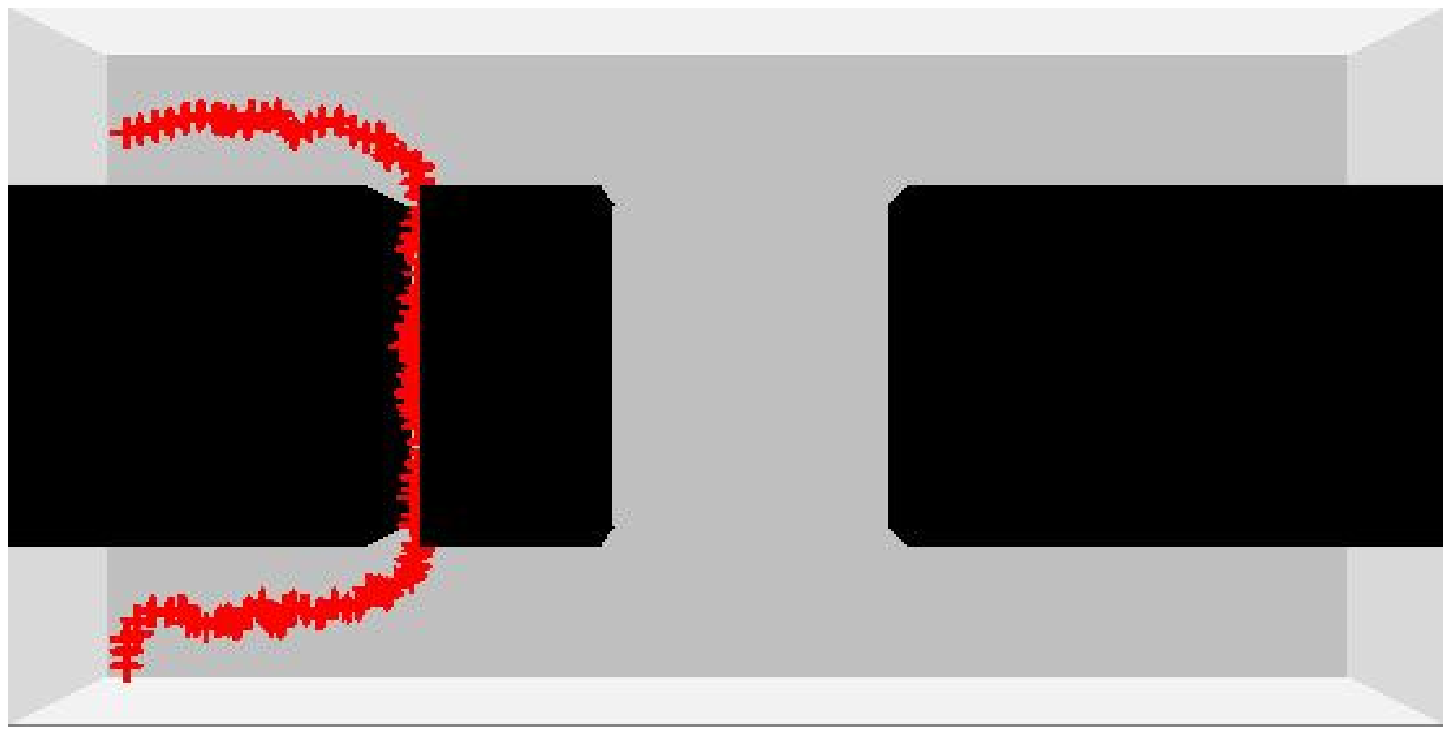}
  \caption{DR-RRT}
  \label{fig:drrrt_boxes}
\end{subfigure}
  \centering
  \begin{subfigure}{0.45\textwidth}
  \centering
  \includegraphics[width=1\textwidth]{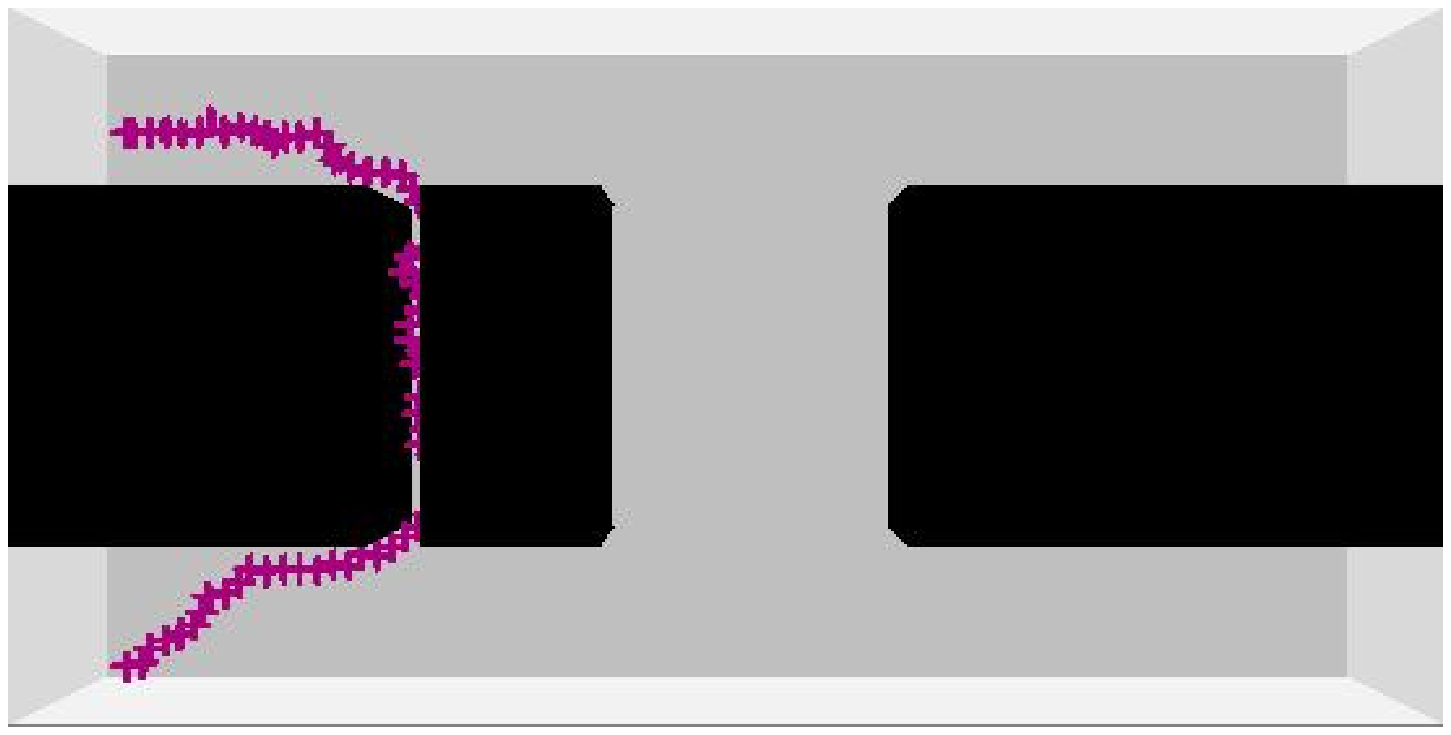}
  \caption{MA-RRT}
  \label{fig:marrt_boxes}
\end{subfigure}
\caption{Five extracted paths from 40 random runs in the Boxes environment.}
\label{fig:boxes_results}
\end{figure*}
\begin{figure*}[htb]
  \centering
  \begin{subfigure}{0.45\textwidth}
  \centering
  \includegraphics[width=1\textwidth]{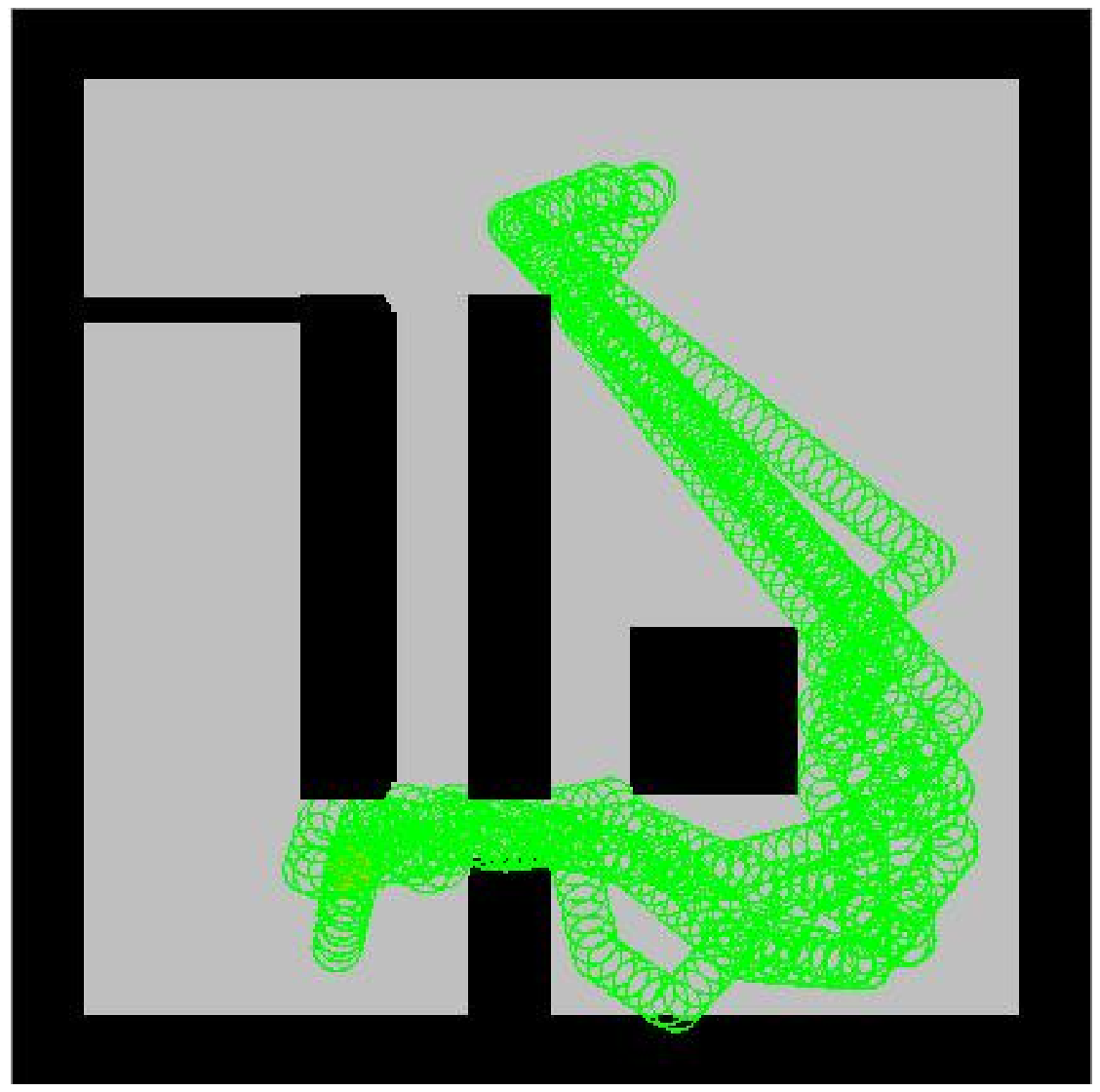}
  \caption{AB-RRT}
  \label{fig:abrrt_walls}
\end{subfigure}
  \centering
  \begin{subfigure}{0.45\textwidth}
  \centering
  \includegraphics[width=1\textwidth]{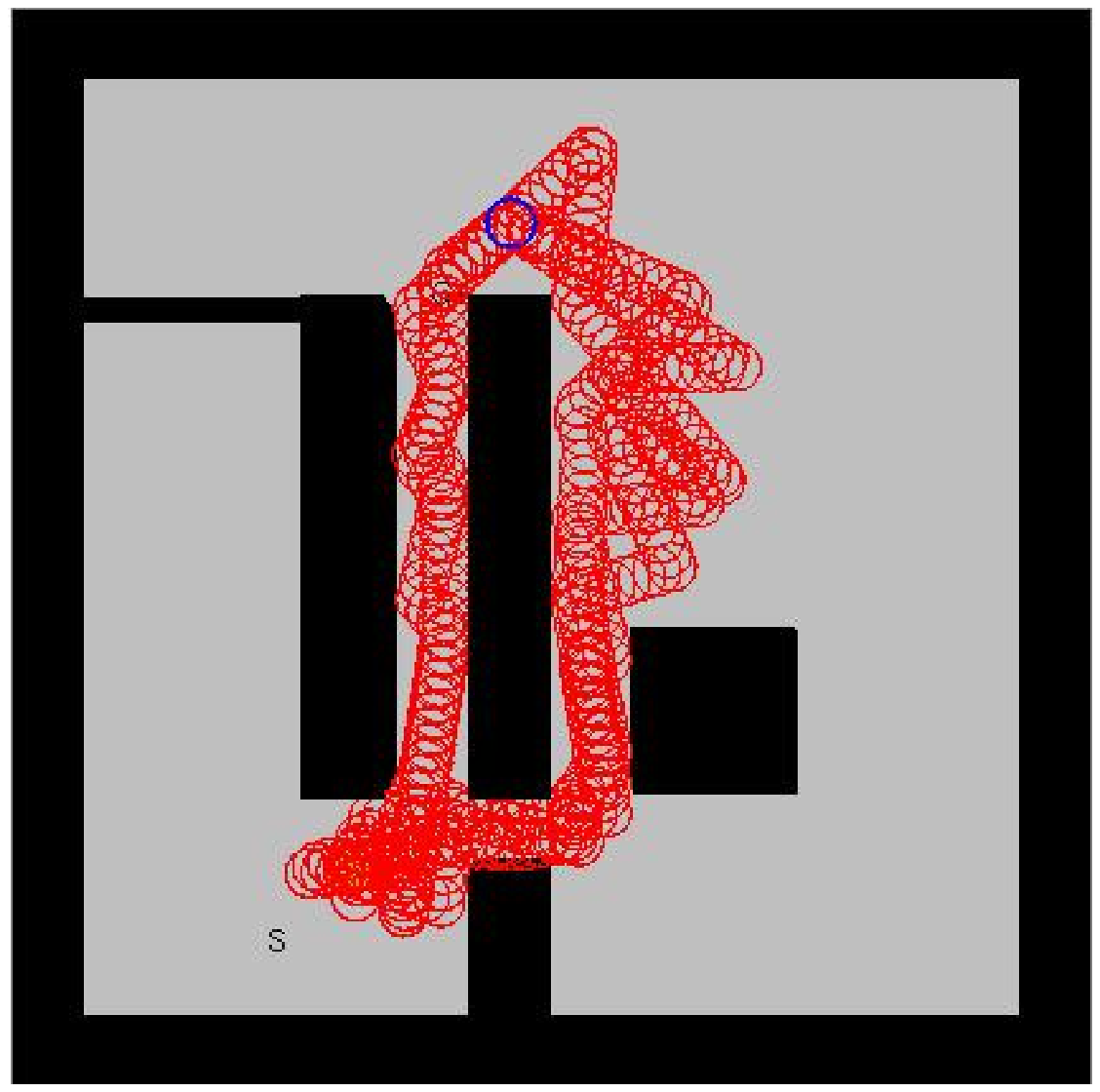}
  \caption{DR-RRT}
  \label{fig:drrrt_walls}
\end{subfigure}
  \centering
  \begin{subfigure}{0.45\textwidth}
  \centering
  \includegraphics[width=1\textwidth]{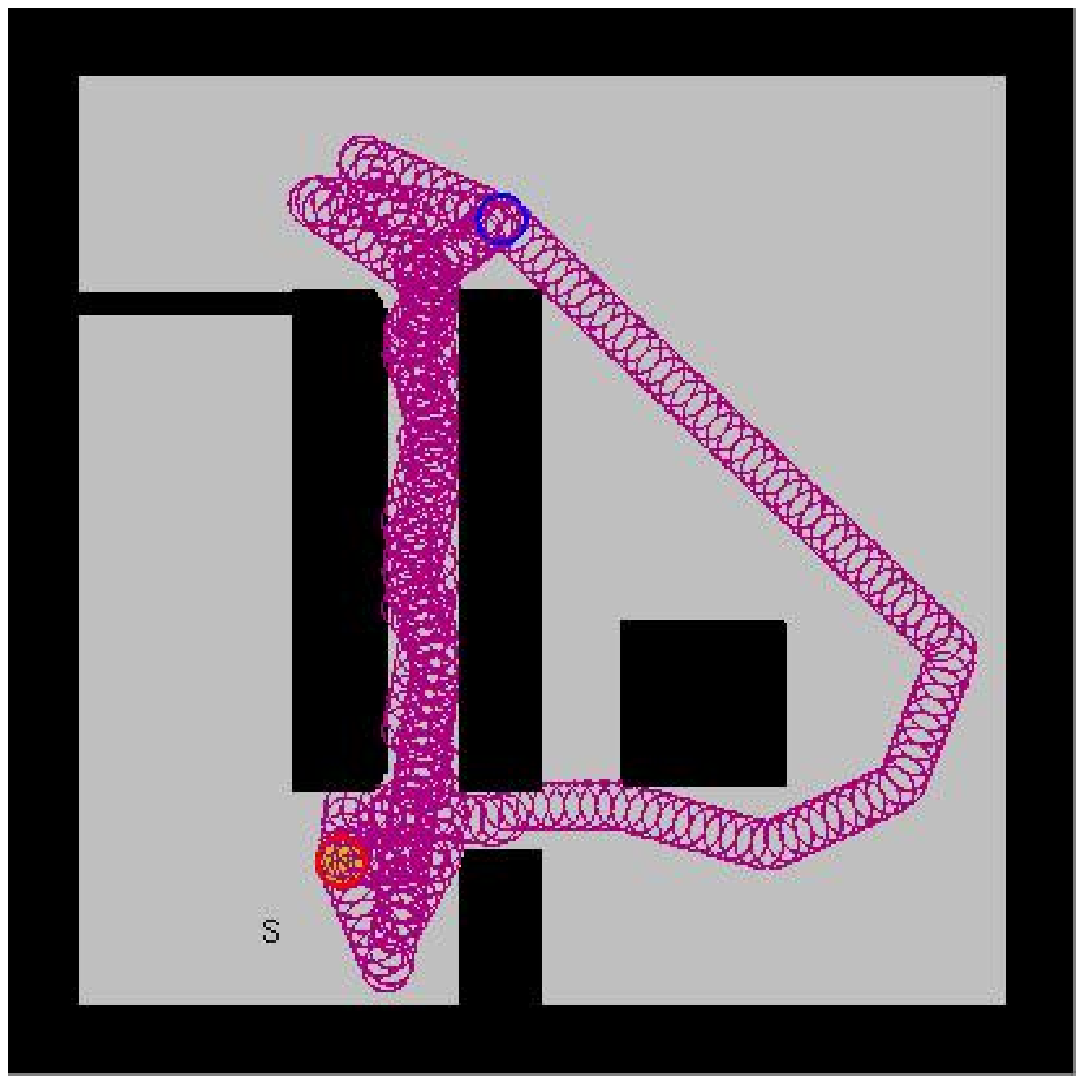}
  \caption{MA-RRT}
  \label{fig:marrt_walls}
\end{subfigure}
\caption{Five extracted paths from 40 random runs in the Walls environment.}
\label{fig:walls_results}
\end{figure*}

Figures \ref{fig:boxes_results} and \ref{fig:walls_results} show
distinct paths returned by each planner.
In both environments, AB-RRT with maximum clearance bias, returned 100\% of its paths passing through
the higher clearance regions of the environment, while DR-RRT and MA-RRT fluctuated
between trials.\par
In the Boxes environment, DR-RRT and MA-RRT returned paths that went through the first
first sequence of regions encountered between the start and the goal.
In the Walls environment, the skeleton guided methods took about the same time
to solve the query. However, AB-RRT returned paths with
higher clearance, which are more likely for the iRobot Create to execute
successfully compared to the shorter paths returned by the other two strategies.
MA-RRT succeeded to return paths through the safe regions 18\% of the time, but it took more than
10 times longer than it took the guided planners.\par

\begin{table*}[htb]
  \caption{Running time (seconds) in robotics environments averaged over 40 runs. \\
           Total planning time reported includes skeleton annotation time.}
\label{tab:runtime}
\begin{center}
\begin{tabular}{|c|r|r|r|r|r|r|r|r|r|r|}
\hline
Strategy & \multicolumn{5}{|c|}{3D Boxes} & \multicolumn{5}{|c|}{Walls} \\
\cline{2-11}
         & \multicolumn{4}{|c|}{Time} & \multicolumn{1}{|c|}{Success}  & \multicolumn{4}{|c|}
                               {Time} & \multicolumn{1}{|c|}{Success} \\ \cline{2-5} \cline{7-10}
         & \multicolumn{2}{|c|}{Skeleton} & \multicolumn{2}{|c|}{Planning} &\multicolumn{1}{|c|}{rate}
         & \multicolumn{2}{|c|}{Skeleton} & \multicolumn{2}{|c|}{Planning} &\multicolumn{1}{|c|}{rate} \\ \cline{2-5} \cline{7-10}
           & avg & $\sigma$ & avg & $\sigma$ & \multicolumn{1}{|c|}{(\%)}& avg & $\sigma$ & avg & $\sigma$ & \multicolumn{1}{|c|}{(\%)}\\ \hline
MA-RRT     & $\NA$ & $\NA$ & 0.94    & 0.36    & 0 &$\NA$ &$\NA$ & 1.99    & 0.01  & 18\\ \hline
DR-RRT     &\textbf{0.22} &0.01 & \textbf{0.66}    & 0.07    & 0  &\textbf{0.004} & 0.001& \textbf{0.043}  & 0.017  &0 \\ \hline
\textbf{AB-RRT} &0.23 &0.02 & 0.72 & 0.03 & \textbf{100} & 0.005 &0.001 & 0.047 & 0.020 &\textbf{100} \\ \hline
\end{tabular}
\end{center}
\end{table*}

\begin{figure*}[htb]
\centering
\begin{subfigure}{0.5\textwidth}
  \centering
\includegraphics[width=1\textwidth]{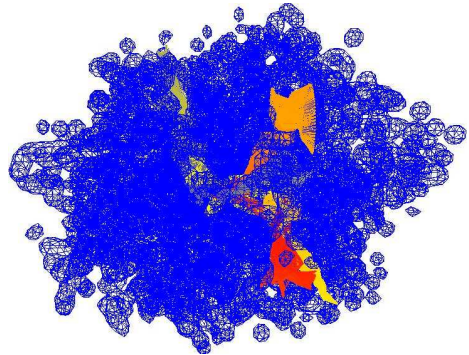}
\caption{DhaA protein and its 5 tunnels}
\label{fig:4hzg_path}
\end{subfigure}

\centering
\begin{subfigure}{0.6\textwidth}
  \centering
\includegraphics[width=1.2\textwidth]{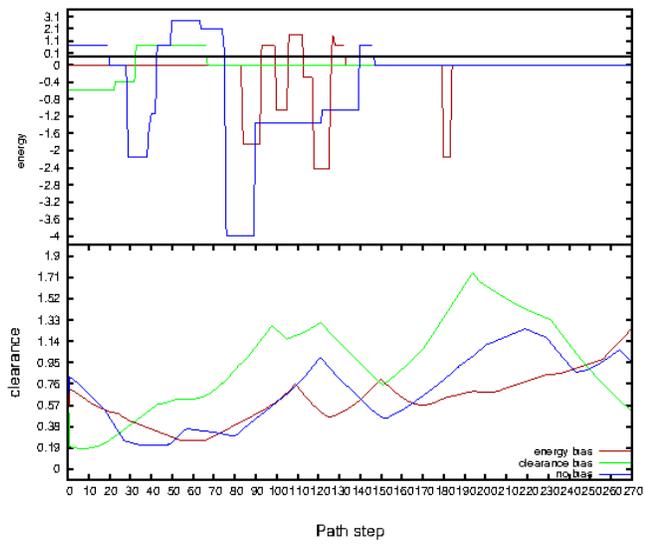}
\caption{DhaA}
\label{fig:4hzg_paths}
\end{subfigure}
\centering
\begin{subfigure}{0.6\textwidth}
  \centering
\includegraphics[width=1.2\textwidth]{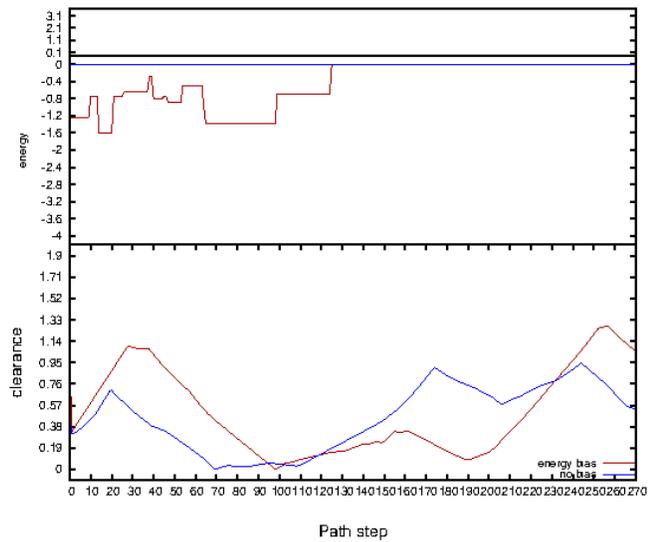}
\caption{DhaA31}
\label{fig:3rk4_paths}
\end{subfigure}
\caption{Best paths found by DR-RRG and AB-RRG in DhaA and DhaA31.}
\label{fig:DhaA_profile}
\end{figure*}

Table \ref{tab:bioruntime} and Figure \ref{fig:DhaA_profile} show the running
time and path profiles for DhaA and its variant DhaA31.
Across the board, DhaA31 paths were found in less time compared to the wild type case.
Running time correlates with binding site accessibility, signaling that the
easier a tunnel is to plan through, the easier it could be for the ligand to
traverse it.\par
In DhaA wild type, AB-RRG with energy bias returned predicted paths in less time than DR-RRG,
showing the advantage of biasing in problems with complex geometry. Moreover,
AB-RRG with energy bias returned all five expected tunnels  in less time
than it took DR-RRG to discover two. Figure \ref{fig:4hzg_paths} shows that high clearance is not always
correlated with low energy. This is motivation for our future plan to study how multiple
metrics can be combined to find an optimal path balancing them.\par
In DhaA31, AB-RRG with clearance bias did not find any paths in the limited time
set of 3 minutes. This is in agreement with the fact that DhaA31 was modified by
reducing its access tunnels volume to increase stability of the reaction with
TCP.
Figure \ref{fig:3rk4_paths} shows that AB-RRG correctly identified a path
through the energetically favorable tunnel.
Using AB-RRG, we are able to identify both tunnels present in the protein and get an intuition on which protein variant is more accessible by looking at the time it took to identify those tunnels.\par
These results show that motion planning algorithms can be reliably used to study the accessibility of protein binding sites. Moreover, the guidance of a skeleton annotated with energy information allows fast discovery of relevant tunnels.\par

\begin{table*}[htb]
  \caption{Running time (seconds) in protein environments averaged over 10 runs. \\
           Success rate = Tunnels discovered / Tunnels expected. }
  \label{tab:bioruntime}
\begin{center}
\begin{tabular}{|c|c|r|r|r|r|r|r|r|r|r|r|}
\hline
Strategy & Bias & \multicolumn{5}{|c|}{DhaA} & \multicolumn{5}{|c|}{DhaA31} \\
\cline{3-12}
& & \multicolumn{4}{|c|}{Time} &\multicolumn{1}{|c|}{Success} & \multicolumn{4}{|c|}{Time} & \multicolumn{1}{|c|}{Success} \\ \cline{3-6} \cline{8-11}
& & \multicolumn{2}{|c|}{Skeleton} & \multicolumn{2}{|c|}{Planning} &\multicolumn{1}{|c|}{rate} &\multicolumn{2}{|c|}{Skeleton} & \multicolumn{2}{|c|}{Planning} &\multicolumn{1}{|c|}{rate} \\ \cline{3-6} \cline{8-11}
& & avg & $\sigma$ & avg & std &\multicolumn{1}{|c|}{(\%)} & avg & $\sigma$ &
avg & $\sigma$ &\multicolumn{1}{|c|}{(\%)} \\ \hline
DR-RRG   & $\NA$ & \textbf{27.5}  & 0.5 & 239.7  & 14.7 & 40 & \textbf{30.5} & 0.3 &202.3&24.5&50 \\ \cline{2-12}
        & energy   & $\NA$ &  $\NA$ &  $\NA$ & $\NA$ & $\NA$ & $\NA$ & $\NA$ & $\NA$ & $\NA$ & $\NA$\\ \hline
\textbf{AB-RRG}   & clearance & 27.8  & 0.2 & 177.9 & 0.3 & 40 &  32.6 & 0.5 &180.0 &0.0 & - \\ \cline{2-12}
         & energy & 27.8  & 0.2 & \textbf{176.0} & 6.0 & \textbf{100} & 32.6 & 0.5 & \textbf{156.4} &16.3 & \textbf{100}\\ \hline
\end{tabular}
\end{center}
\end{table*}

\section{Conclusion}
We presented an annotated-skeleton biased planning strategy to find
environment-informed guided paths for motion planning problems.
By separating biasing from the planning process, we were able to use different
planning strategies to solve a variety of problems.
We showed the merit of the AB-RRT method in robotics problems where obstacle clearance is important.
For protein-ligand binding problems, we used AB-RRG to assess the accessibility of protein binding sites and find feasible paths
through protein tunnels.
In the future, we plan to study how multiple metrics can be combined to
dynamically discover relevant regions to a specific query.
We also plan to use the annotated skeleton in dynamic environments
where the topology of the problem stays the same and a planner needs guidance
based on changes in the environment.


\clearpage
\bibliographystyle{splncs03}
\bibliography{annotated_skeleton_planning.bbl}

\begin{thebibliography}{10}
\providecommand{\url}[1]{\texttt{#1}}
\providecommand{\urlprefix}{URL }

\bibitem{abdjv-obprm-98}
Amato, N.M., Bayazit, O.B., Dale, L.K., Jones, C., Vallejo, D.: {OBPRM}: an
  obstacle-based {PRM} for 3d workspaces. In: Proceedings of the third Workshop
  on the Algorithmic Foundations of Robotics. pp. 155--168. A. K. Peters, Ltd.,
  Natick, MA, USA (1998), (WAFR `98)

\bibitem{bsa-lbobprm-01-book}
Bayazit, O.B., Song, G., Amato, N.M.: Ligand binding with obprm and haptic user
  input: Enhancing automatic motion planning with virtual touch. In:
  El-Mabrouk, N., Lengauer, T., Sankoff, D. (eds.) Currents in Computational
  Molecular Biology. pp. 81--82. Les Publications CRM, Montreal, Canada (2001),
  book includes short papers from The Fifth ACM International Conference on
  Computational Molecular Biology ({\em RECOMB}), Montreal, Canada, April 2001.

\bibitem{blk-tcsrp-2012}
Bhattacharya, S., Likhachev, M., Kumar, V.: Topological constraints in
  search-based robot path planning. Autonomous Robots  33(3) (2012)

\bibitem{c-edt-92}
Chiang, C.S.: The {Euclidean} distance transform. Ph.{D}. thesis, Dept. Comput.
  Sci., Purdue Univ., West Lafayette, IN (Aug 1992), report CSD-TR 92-050

\bibitem{dsba-drbrrt-16}
Denny, J., Sandstrom, R., Bregger, A., Amato, N.M.: Dynamic region-biased
  exploring random trees. In: Proc.\ Int.\ Workshop on Algorithmic Foundations
  of Robotics ({WAFR}). San Francisco, CA (December 2016)

\bibitem{hpc-tpiasocce-2019}
Ha, J.S., Park, S.S., Choi, H.L.: Topology-guided path integral approach for
  stochastic optimal control in cluttered environment. Robotics and Autonomous
  Systems  113,  81 -- 93 (2019),
  \url{http://www.sciencedirect.com/science/article/pii/S0921889017308874}

\bibitem{hk-fuwmapp-00}
Holleman, C., Kavraki, L.E.: A framework for using the workspace medial axis in
  prm planners. In: Proc.\ {IEEE} Int.\ Conf.\ Robot.\ Autom.\ ({ICRA}).
  vol.~2, pp. 1408--1413. San Franasisco, CA (2000)

\bibitem{vdmt-trmpgdei-2013}
Ivan, V., Zarubin, D., Toussaint, M., Komura, T., Vijayakumar, S.:
  Topology-based representations for motion planning and generalization in
  dynamic environments with interactions. The International Journal of Robotics
  Research  32(9-10),  1151--1163 (2013),
  \url{https://doi.org/10.1177/0278364913482017}

\bibitem{r-rrgmpmmr}
Kala, R.: {Rapidly exploring random graphs: motion planning of multiple mobile
  robots}. Advanced Robotics  27(14),  1113--1122 (2013),
  \url{https://doi.org/10.1080/01691864.2013.805472}

\bibitem{kaushik2018}
Kaushik, S., Marques, S.M., Khirsariya, P., Paruch, K., Libichova, L.,
  Brezovsky, J., Prokop, Z., Chaloupkova, R., Damborsky, J.: Impact of the
  access tunnel engineering on catalysis is strictly ligand specific.
  Federation of European Biochemical Societies Journal  285(8),  1456--1476
  (2018)

\bibitem{kslo-prpp-96}
Kavraki, L.E., \v{S}vestka, P., Latombe, J.C., Overmars, M.H.: Probabilistic
  roadmaps for path planning in high-dimensional configuration spaces. {IEEE}
  Trans.\ Robot.\ Automat.  12(4),  566--580 (August 1996)

\bibitem{KLVANA20091339}
Klvana, M., Pavlova, M., Koudelakova, T., Chaloupkova, R., Dvorak, P., Prokop,
  Z., Stsiapanava, A., Kuty, M., Kuta-Smatanova, I., Dohnalek, J., Kulhanek,
  P., Wade, R.C., Damborsky, J.: Pathways and mechanisms for product release in
  the engineered haloalkane dehalogenases explored using classical and random
  acceleration molecular dynamics simulations. Journal of Molecular Biology
  392(5),  1339 -- 1356 (2009),
  \url{http://www.sciencedirect.com/science/article/pii/S0022283609008092}

\bibitem{Lahoda:kw5074}
Lahoda, M., Mesters, J.R., Stsiapanava, A., Chaloupkova, R., Kuty, M.,
  Damborsky, J., {Kuta Smatanova}, I.: {Crystallographic analysis of
  1,2,3-trichloropropane biodegradation by the haloalkane dehalogenase DhaA31}.
  Acta Crystallographica Section D  70(2),  209--217 (Feb 2014),
  \url{https://doi.org/10.1107/S1399004713026254}

\bibitem{l-mpjrmdaoa-99}
Latombe, J.C.: Motion planning: A journey of robots, molecules, digital actors,
  and other artifacts. Int.\ J.\ Robot.\ Res.  18(11),  1119--1128 (1999)

\bibitem{lj-qemppbs-05}
Laurie, A.T., Jackson, R.M.: Q-sitefinder: an energy-based method for the
  prediction of protein-ligand binding sites  21(9),  1908--1916 (2005)

\bibitem{lk-rkp-01}
{LaValle}, S.M., Kuffner, J.J.: Randomized kinodynamic planning. Int.\ J.\
  Robot.\ Res.  20(5),  378--400 (May 2001)

\bibitem{lta-gfsmafs-03}
Lien, J.M., Thomas, S., Amato, N.: A general framework for sampling on the
  medial axis of the free space. In: Proc.\ {IEEE} Int.\ Conf.\ Robot.\ Autom.\
  ({ICRA}). vol.~3, pp. 4439--4444 (sept 2003)

\bibitem{chimera2004}
Pettersen, E.F., Goddard, T.D., Huang, C.C., Couch, G.S., Greenblatt, D.M.,
  Meng, E.C., Ferrin, T.E.: Ucsf chimera: A visualization system for
  exploratory research and analysis. Journal of Computational Chemistry
  25(13),  1605--16012 (2004)

\bibitem{PDB-webpage}
The {Protein Data Bank}, {\tt http://www.rcsb.org/pdb/}

\bibitem{r-cpmpg-79}
Reif, J.H.: Complexity of the mover's problem and generalizations. In: Proc.\
  {IEEE} Symp.\ Foundations of Computer Science ({FOCS}). pp. 421--427. San
  Juan, Puerto Rico (October 1979)

\bibitem{rhb-iotpodt-2017}
Rösmann, C., Hoffmann, F., Bertram, T.: Integrated online trajectory planning
  and optimization in distinctive topologies. Robotics and Autonomous Systems
  88,  142 -- 153 (2017),
  \url{http://www.sciencedirect.com/science/article/pii/S0921889016300495}

\bibitem{rmft-otppsrn-2017}
Rösmann, C., Oeljeklaus, M., Hoffmann, F., Bertram, T.: Online trajectory
  prediction and planning for social robot navigation. pp. 1255--1260 (07 2017)

\bibitem{taos-mcs-12}
Tagliasacchi, A., Alhashim, I., Olson, M., Zhang, H.: Mean curvature skeletons.
  Eurographics Symposium on Geometry Processing 2012  27(1) (2012)

\bibitem{vk-tdpssmp-2017}
Von{\'a}sek, V., Kozl{\'i}kov{\'a}, B.: Tunnel detection in protein structures
  using sampling-based motion planning. In: 2017 11th International Workshop on
  Robot Motion and Control (RoMoCo). pp. 185--192 (July 2017)

\bibitem{yl-fpfmsbrng-00}
Yang, L., {LaValle}, S.M.: A framework for planning feedback motion strategies
  based on a random neighborhood graph. In: Proc.\ {IEEE} Int.\ Conf.\ Robot.\
  Autom.\ ({ICRA}). pp. 554--549 (2000)

\bibitem{yb-asdppbama-04}
Yang, Y., Brock, O.: Adapting the sampling distribution in prm planners based
  on an approximated medial axis. In: Proc.\ {IEEE} Int.\ Conf.\ Robot.\
  Autom.\ ({ICRA}). vol.~5, pp. 4405--4410 (2004)

\bibitem{ytea-uudobp-12}
Yeh, H.Y., Thomas, S.L., Eppstein, D., Amato, N.M.: {UOBPRM:} {A} uniformly
  distributed obstacle-based {PRM}. In: Proc.\ {IEEE} Int.\ Conf.\ Intel.\
  Rob.\ Syst.\ ({IROS}). pp. 2655--2662 (2012),
  \url{http://dx.doi.org/10.1109/IROS.2012.6385875}

\end{thebibliography}

\end{document}